\documentclass[11pt]{article}

\usepackage[final]{acl}
\usepackage{nert}
\usepackage{hyperref}
\usepackage{graphicx}
\usepackage{subcaption}

\usepackage{times}
\usepackage{latexsym}

\usepackage[T1]{fontenc}

\usepackage[utf8]{inputenc}

\usepackage{microtype}

\usepackage{inconsolata}

\usepackage{graphicx}
\usepackage{linguex}
\usepackage{tabularx}
\usepackage{dsfont}

\definecolor{customred}{HTML}{9e2a2b}
\definecolor{customblue}{HTML}{2D43BF}
\definecolor{bright_red}{HTML}{bd5363}
\definecolor{bright_green}{HTML}{52c755}
\definecolor{yellow}{HTML}{b6b83d}
\definecolor{purple}{HTML}{b35fd4}
\definecolor{orange}{HTML}{6a8bd9}

\renewcommand{\nertcomment}[4]{\unskip}
\title{Language Models Learn Constructional Semantics, \textit{Not To Mention} Syntax: Investigating LM Understanding of \pf  Constructions}

\author{Wesley Scivetti\textsuperscript{1}
\quad Ethan Wilcox\textsuperscript{1} \quad Nathan Schneider\textsuperscript{1}\\\textbf{Kanishka Misra}\textsuperscript{2} \quad \textbf{Leonie Weissweiler}\textsuperscript{3} \\
  \textsuperscript{1}Georgetown University\textsuperscript{1}, \textsuperscript{2}The University of Texas at Austin\\\textsuperscript{3}Leipzig University \& ScaDS.AI Dresden/Leipzig \\
  \emldisplay{wss37@georgetown.edu}{wss37@georgetown.edu}
   \\
}

\begin{document}
\maketitle
\begin{abstract}

Grasping the semantics of rare \textit{constructions} (form--meaning pairings) has been shown to be a challenging problem that has currently only been solved by the largest LLMs. It remains an open question if open-source models have robust constructional understanding, and if so, what learning dynamics underlie the acquisition of this knowledge. Focusing on a set of rare \pf constructions in English (e.g.~``let alone'', ``much less''), we construct a novel dataset to test their meanings using both scalar adjectival semantics and general world knowledge. Testing a wide range of models differing in parameter count, architecture, and pretraining dataset size, we find that several modestly sized models are sensitive to both the forms and the meanings of \pf constructions, though models trained on human-scale data fail at all meaning evaluations. Turning to training dynamics for a set of open-checkpoint models, we find that \pf understanding emerges later in training than \pf syntactic knowledge, and that learning of \pf semantics is correlated with gains in some domains of world knowledge. Overall, our empirical results support the conclusion that modestly sized open-source models can grasp the rare \pf constructions, and demonstrate a connection between knowledge of \pf constructions and other meaning domains.\footnote{\url{https://github.com/WesScivetti/Meaning\_Alone}}

\end{abstract}

\section{Introduction}

\interfootnotelinepenalty=10000

Language models (LMs) have great potential as tools for studying language. The success of a domain-general statistical learner on many linguistic tasks has prompted some researchers to argue that LMs should be more central in evaluating and developing linguistic theory \citep{warstadt2022artificial,Futrell_Mahowald_2025}. In this work, we focus on constructionist approaches, or Construction Grammar (\citealp{Goldberg_1995,Goldberg_2006,Croft_2001}, \textit{inter alia}), a family of linguistic theories which posit that language is primarily structured as form-function mappings of gradient complexity. Construction Grammar accounts for not only the most integral parts of linguistic structure (e.g.\ verb argument structures; \citealp{Goldberg_1995}), but also rare phenomena traditionally relegated to the ``periphery'', arguing that both can be accounted for with similar representations \citep{fillmoreRegularityIdiomaticityGrammatical1988}. Construction Grammar’s ability to describe and account for rare linguistic phenomena makes it a valuable framework for studying language models, as mastery of rare and complex constructions is a crucial part of humans' linguistic knowledge, and may be particularly challenging for language models due to scarcity in their input.

At the same time, the general success of neural language models, which do not explicitly distinguish between syntax and semantic information, has led some researchers to argue that Construction Grammar is more aligned with LM processing of language relative to other frameworks \citep{weissweiler-etal-2023-construction,Goldberg_2024,Piantadosi_2024}. In order to fully evaluate these claims, it is worth investigating the extent to which LMs can serve as ``model learners'' \citep{warstadt2022artificial} of constructionist theories of language.

\begin{table*}[]

\centering\small
\begin{tabular}{lrr|rrr}
\toprule
\textbf{Cxn}  & \textbf{Olmo3} & \textbf{Pythia} &  \textbf{COCA}  & \textbf{Prop.} & \textbf{Cxn Freq. COCA}  \\
\midrule
\la       &     4261   &  2874  & 8631    & 1.00                 & 8311--8631           \\
\ml        &      8671     &  5521 & 13184 & 0.44                & 4570--7089           \\
\ntm        &  3862   &  2183 & 8803  & 0.38                & 2561--4207           \\
\nvm          &    807    &  796 & 2974   & 0.31                & 677--1209 \\
\bottomrule
\end{tabular}
\caption{\textbf{\pf construction frequencies in LM pretraining data and COCA.} String frequencies are reported for Olmo3 pretraining data, Pythia pretraining data (The Pile) and COCA. All counts are normalized per billion tokens. Prop. refers to the proportion of strings in COCA that were true instances of the construction out of 100 sampled instances. Cxn Freq. refers to the approximate frequency of the construction based on 95\% confidence intervals of the sampled proportion in COCA. LM pretraining counts computed using infini-gram \citep{Liu_Min_Zettlemoyer_Choi_Hajishirzi_2024}.}
\label{tab:cxn_freqs}
\end{table*}

Because Construction Grammar posits that form and meaning are not separable parts of linguistic knowledge, a constructionist ``model system'' should learn and understand constructions with respect to both their forms and functions. However, up to this point, there are somewhat mixed results regarding LMs' syntactic and semantic knowledge of rare constructions. There is ample evidence that even relatively small language models learn formal properties of rare constructions, including \textsc{Article-Adjective-Numeral-Noun} \citep{misraLanguageModelsLearn2024}, \la \citep{rozner-etal-2025-babylms,scivetti-etal-2025-unpacking}, and the \textsc{Comparative-Correlative} \citep{Weissweiler_Hofmann_Köksal_Schütze_2022}. However, in many cases, these smaller models seem unable to grasp the semantics of these rare constructions \citep{Weissweiler_Hofmann_Köksal_Schütze_2022,scivetti-etal-2025-unpacking}. On the other hand, extremely large LLMs have been shown to have relatively nuanced semantic understanding of a range of constructions \citep{mortensen-etal-2024-verbing,scivetti-etal-2025-beyond}. Since most LLM evaluation has been done in the prompting setting on closed-source LLMs, it remains unclear whether semantic understanding of constructions can be observed in the raw probabilities of (smaller) language models, particularly for exceedingly rare constructions.


Additionally, it is not clear that all current evaluation datasets are well suited for testing constructional semantic understanding. Past work relies on unnatural sounding test items \citep{Weissweiler_Hofmann_Köksal_Schütze_2022,scivetti-etal-2025-unpacking} or more complex metalinguistic tasks which may limit LM performance \citep{bonialConstructionGrammarCorpus2024b}. Additionally, particularly for rare constructions, it is likely that constraints on their behavior are related to more general phenomena in language \citep{Potts_2024,misraLanguageModelsLearn2024}, and it is reasonable to expect that the semantics of constructions are shaped and constrained by their interactions with other domains of semantics.

In this work, we address the above gaps with a new dataset for testing a rare\footnote{For approximate corpus frequencies see Table \ref{tab:cxn_freqs}.} family of four constructions: \la, \ml, \ntm, and \nvm, collectively known as \pf constructions because they conjoin two phrases that are both in focus:

\ex.\label{ex:pf} He doesn't like shrimp, let alone squid. \citep{fillmoreRegularityIdiomaticityGrammatical1988}

We evaluate \pf syntax and semantics on a much wider range of models than has been tested in past work on constructional semantics.\nss{give some brief insight into why the term ``paired focus'' and why such constructions are interesting} We ground the creation of our dataset in other factors in language production which are likely to inform learning of the constructions' semantics: scalar adjectives and general world knowledge.

This allows us to ask and answer detailed research questions about the acquisition of \pf semantics.  

\textbf{1) How do training data, parameter count, and pretraining objective impact knowledge of \pf meaning?}
We show that a range of medium-sized open-source models show sensitivity to the construction-level semantics in their raw probability distributions. However, we do not observe any above-chance knowledge of \pf semantics for models trained on human-scale data (BabyLMs, \citealp{warstadtFindingsBabyLMChallenge2023}).
\textbf{2) When are \pf form and meaning learned throughout pretraining?}
We show that \pf form is consistently acquired prior to \pf semantics. Furthermore, we show that inferences early in learning are often influenced by typicality (world knowledge) rather than constructional meaning, particularly for weaker models. We find that the different \pf constructions we test are highly correlated with one another in terms of both syntactic and semantic performance.
\textbf{3) How does \pf meaning correlate with performance on other linguistic benchmarks?}
We examine the learning dynamics of both the form and meaning of the \pf constructions, in relation to performance on a range of existing linguistic benchmarks. We find that performance on our syntactic evaluations of the constructions plateaus early in pretraining (mirroring learning curves for the BLiMP \citep{warstadtBLiMPBenchmarkLinguistic2020b} grammatical benchmark), while functional knowledge is acquired much later and more closely mirrors the learning trajectory of the world-knowledge-based EWoK \citep{ivanova-etal-2025-elements} benchmark.

Overall, our results show that \pf constructions can be learned by relatively small models (under 400M parameters). They also shed light on the interrelatedness of \pf constructions with scalar semantics, and with domains of world knowledge in which such scalar semantics are relevant. Furthermore, the failure of all ``human-scale'' models on our semantic evaluation points calls into question the ability of human-scale pure text LMs to serve as model systems of linguistic theories that posit joint acquisition of form and meaning.

\section{Background}
\subsection{Paired-Focus Constructions}

\begin{table*}[]
\centering\small\smaller\setlength{\tabcolsep}{2pt}
\begin{tabular}{@{}rllllll@{}}
\toprule
Idx &Name/Description        &   Feat. & Sent.1   & Feat. & Sent. 2: \textit{\_\_\_ than lifting a huge one.}  & Pairwise Feat.      \\
\midrule
6&Cxn Entails Plaus. & \textcolor{purple}{\textbf{+Cxn}}&I couldn't lift a tiny rock, \textcolor{purple}{\textbf{let alone}} a huge one. & \textcolor{bright_green}{\textbf{+Plaus.}}&Lifting a tiny rock is \textbf{easier} 
& \textcolor{bright_green}{\textbf{Entailment}}    \\
7&Cxn Contradicts Implaus. & \textcolor{purple}{\textbf{+Cxn}}&I couldn't lift a tiny rock, \textcolor{purple}{\textbf{let alone}} a huge one. & \textcolor{bright_red}{\textbf{$-$Plaus.}}&Lifting a tiny rock is \textbf{harder} 
& \textcolor{bright_red}{\textbf{Contradiction}} \\
\midrule
8&Neutral Plaus.\ Control & \textcolor{yellow}{\textbf{$-$Cxn}}&I couldn't lift a tiny rock, \textcolor{yellow}{\textbf{or}} a huge one.             & \textcolor{bright_green}{\textbf{+Plaus.}}&Lifting a tiny rock is \textbf{easier} 
& \textcolor{orange}{\textbf{Neutral}}      \\
9&Neutral Implaus.\ Control & \textcolor{yellow}{\textbf{$-$Cxn}}&I couldn't lift a tiny rock, \textcolor{yellow}{\textbf{or}} a huge one.             & \textcolor{bright_red}{\textbf{$-$Plaus.}}&Lifting a tiny rock is \textbf{harder} 
& \textcolor{orange}{\textbf{Neutral}}       \\
\midrule
10&Cxn Contradicts Plaus. & \textcolor{purple}{\textbf{+Cxn}}& I couldn't lift a huge rock, \textcolor{purple}{\textbf{let alone}} a tiny one.   & \textcolor{bright_green}{\textbf{+Plaus.}}&Lifting a tiny rock is \textbf{easier} 
& \textcolor{bright_red}{\textbf{Contradiction}} \\
11&Cxn Entails Implaus. & \textcolor{purple}{\textbf{+Cxn}}&I couldn't lift a huge rock, \textcolor{purple}{\textbf{let alone}} a tiny one.   & \textcolor{bright_red}{\textbf{$-$Plaus.}}&Lifting a tiny rock is \textbf{harder} 
& \textcolor{bright_green}{\textbf{Entailment}}  \\ 
\bottomrule
\end{tabular}

\caption{\pf semantics dataset overview. Sent.~1 serves as the context and Sent.~2 as the target.
}
\label{tab:examples}
\end{table*}

In this work, we focus on four related \pf constructions: \la, \ml, \nvm, and \ntm. We focus on these constructions because their syntactic and semantic properties are well established in Construction Grammar theory \citep{fillmoreRegularityIdiomaticityGrammatical1988}, and have also been studied in past computational work (e.g.\ \citealp{bonialConstructionGrammarCorpus2024b,rozner-etal-2025-constructions,rozner-etal-2025-babylms,scivetti-etal-2025-unpacking}), though past work on investigating language model understanding of \pf semantics has yielded mostly negative results. 

\citet{fillmoreRegularityIdiomaticityGrammatical1988} provide a comprehensive treatment of the \la construction, though much of their analysis can be extended to the other \pf constructions in this work. Syntactically, these \pf constructions function somewhat similarly to coordinating conjunctions (Examples \ref{ex:pf} and \ref{ex:and}), but resist various types of syntactic movement (Example \ref{ex:movement}), and generally behave like negative polarity items (Example \ref{ex:npi}).

\ex.\label{ex:and} He doesn't like shrimp, or squid. 

\ex.\label{ex:movement} ??It's shrimp, not to mention squid, that he doesn't like.

\ex.\label{ex:npi} ??I like shrimp, much less squid.

Regarding \pf semantics, the constructions indicate a relationship between two conjoined and focused elements which are being compared. The comparison between the focused elements evokes a scalar relationship with the two elements representing ``two
points on a scale'' \citep{fillmoreRegularityIdiomaticityGrammatical1988}. Generally, the second focused element has a higher value on the scale than the first focused element. Overall, the semantics of \pf constructions are related to scalar semantics more generally (in the invocation of a scale) and to world knowledge, which informs what scalar properties are natural for the focused items in the construction.



\subsection{Related Work}\label{sec:related}

\subsubsection{Linguistic Capabilities of LMs}

Broadly, this work follows in a long line of research seeking to investigate LM knowledge of linguistic phenomena by examining LM probabilities for grammatical and ungrammatical sequences (\citealp{hu-etal-2020-systematic}, \textit{inter alia}). Among the most relevant of these works are those which seek to design datasets to assess broad domains of linguistic abilities, whether that be syntactic, conceptual, or world knowledge. We leverage several of these datasets as comparison points for learning dynamics of \pf constructions. Furthermore, investigations of scalar semantics of adjectives \citep{gari-soler-apidianaki-2020-bert,schuster-etal-2020-harnessing,lin-etal-2024-probing,nizamani-etal-2024-siga} and adverbs \citep{lorge-pierrehumbert-2023-wacky} are of particular relevance to this present work, though most past work has focused on scalar implicature and relative intensity of adjectives, which is orthogonal to the present study. Contributing to this line of work are several datasets containing scalar adjectives of varying intensities \citep{deMelo_Bansal_2013,wilkinson-tim-2016-gold,cocos-etal-2018-learning}. We rely on the dataset of \citet{wilkinson-tim-2016-gold} for the scales and adjectives in our dataset.

\subsubsection{LM Understanding of Rare Constructions}

Past work on evaluating LM understanding of constructions has primarily focused on evaluating various constructions that have been outlined in theoretical linguistics, such as argument structure constructions \citep{Li_Zhu_Thomas_Rudzicz_Xu_2022,Veenboer_Bloem_2023,sung-kyle-2024-leveraging}, the \textsc{Comparative-Correlative} \citep{Weissweiler_Hofmann_Köksal_Schütze_2022}, \textsc{Causal-Excess} \citep{zhou-etal-2024-constructions}, \textsc{NPN} \citep{scivetti-schneider-2025-construction}, \textsc{PiPP} \citep{Potts_2024}, \textsc{AANN} \citep{mahowaldDiscerningSeveralThousand2023c,chronis-etal-2023-method,misraLanguageModelsLearn2024}, and \la \citep{scivetti-etal-2025-unpacking}. Other work opts for a more general investigation of a range of constructions \citep{Tayyar_Madabushi_Romain_Divjak_Milin_2020, 
rozner-etal-2025-babylms,scivetti-etal-2025-beyond}, or discovers constructions using unsupervised methods \citep{Dunn_2017,beuls-van-eecke-2024-humans,verheyen-etal-2025-shall}. While most work on constructional understanding in LMs has focused on English, there is growing work on evaluation of constructions in languages beyond English \citep{Tseng_Shih_Chen_Chou_Ku_Hsieh_2022,bunzeck-etal-2025-construction,huang-etal-2025-assessing,yang-2025-language}. 

\pf constructions have been addressed in several past works. \citet{rozner-etal-2025-constructions} and \citet{rozner-etal-2025-babylms} use Global Affinity measures to show that both RoBERTa and a range of BabyLMs have knowledge of collocations for the \pf constructions \la and \ml. Our results robustly replicate theirs. \citet{bonialConstructionGrammarCorpus2024b} and \citet{scivetti-etal-2025-beyond} include \la as one of several constructions and find that large, closed-source LLMs can perform a metalinguistic grouping task and natural language inference successfully on examples of these constructions. Our work diverges from these approaches in that we test smaller models using direct, probability-based evaluations. 

Our work is most similar in approach to \citet{scivetti-etal-2025-unpacking}, who design probability-based measures for syntactic and semantic properties of \la. As discussed previously, we argue that their dataset for semantics relies on arbitrary contrasts which do not clearly implicate scalar semantics. Furthermore, they only test a single model architecture (OPT-125M) and data scenario (100M tokens of BabyLM), and only address \la and not the highly related \ml, \ntm, and \nvm.

\section{Paired-Focus Dataset}

We construct a novel dataset to test the semantics of the \pf constructions. Past work \citep{scivetti-etal-2025-unpacking} used a fully arbitrary dataset, where there was no obvious scalar relationship between the focused elements (outside of what may be communicated by the construction itself):

\ex.\label{ex:arbitrary} I couldn't lift the orange crate, let alone the green crate. \citep{scivetti-etal-2025-unpacking}

Failure on fully arbitrary evaluations could be due to several factors, apart from the models truly not understanding the construction. It is also possible that they may be unable to map the arbitrary relationship between the two focused elements onto any scale, which would also lead to a failure to understand the construction overall. If this is the case, we would expect the models to fail to comprehend arbitrary examples (like Example \ref{ex:arbitrary}) but succeed at understanding examples with a coherent scale. 
Furthermore, it is possible that the models have only limited knowledge of the construction, where the constructions can only be interpreted in contexts where the scalar relationship between the focused elements is already obvious due to world knowledge.
We would then expect the models to succeed in cases where there is a clear scalar relationship that makes sense in the context of world knowledge but fail in cases where there is a clear scalar relationship which conflicts with world knowledge.




To distinguish between these possibilities, we construct a new dataset that contains \pf examples where the focused elements have clear scalar relationships between them. We use adjectives and scales from \citet{wilkinson-tim-2016-gold} as a starting point for the focused elements that will be compared within the construction. In all examples, we pair adjectives from their dataset with a set of common nouns that are natural sounding with the chosen scalar adjectives\footnote{\pf constructions can focus other syntactic phrases besides noun phrases. To control for the confound of how syntactic structure could influence understanding, we only place noun phrases in the focused slot of the construction.}. In total, we generate 198 starting templates across 4 unique scales from \citet{wilkinson-tim-2016-gold}, and then fill the resulting templates with a set of appropriate adjectives and nouns, resulting in a dataset of 3.5k example sentence pairs per construction. Example sentences for all scales are shown in Table \ref{tab:frames} in Appendix \ref{sec:appendix_data}.

\subsection{Evaluation of Dataset}
To test model sensitivity to \pf semantics, we measure the effect that the constructions have on model output probabilities on a follow-up sentence that would be entailed by the \pf construction. Because \pf constructions imply that the second focused element is a higher value on the scale than the first focused element \citep{fillmoreRegularityIdiomaticityGrammatical1988}, we include follow-up sentences which either entail or contradict this scalar relationship. Models are expected to prefer the follow-up which reinforces the semantics of the construction to one which contradicts the construction (compare 6 and 7 in Table \ref{tab:examples}). 



\nss{why is it not sufficient to compare model's judgments of  \cref{ex:nonarbitrary} vs.~\cref{ex:nonarbitrary2}? because it could be ignoring the `let alone' part and judging based on the first part of the sentence (plausibility)?}

\nss{I am a bit concerned about judging coherence of \cref{ex:nonarbitrary_fo1} vs.~\cref{ex:nonarbitrary_fo2} because the 2nd one is so implausible. I would expect a human to be confused by the statement ``lifting a tiny rock is harder than lifting an huge one''. I wonder if marking it as a hypothetical would help: ``\textbf{Suppose I tell you that} I couldn't lift a tiny rock, let alone an huge one. \textbf{In this world,} lifting a tiny rock is harder''}


Directly comparing the model probabilities of follow up-sentences is confounded by the logical plausibility of the follow-ups independent of the construction. We expect that, irrespective of the presence of a \pf construction, a good model will assign a higher probability to a follow-up that aligns with world knowledge, all other things being equal. To control for this fact, we compare example pairs with a \pf construction to examples \textit{without} a construction with scalar semantics, specifically the simple conjunction ``or'' (see 8 and 9, Table \ref{tab:examples}). Intuitively, if the model is sensitive to \pf semantics, we expect a model to assign higher probability to a follow-up that aligns with the \pf construction, beyond what is assigned to the follow-up due to world knowledge. 







Concretely, given a base sentence $S_{+{\mathbf{Cxn}}}$ and a follow-up sentence $T_{\pm{\mathbf{Plausible}}}$, we calculate the difference $\Delta P$ as follows. Here, $\iota_{\theta} = - \log p_{\theta}$ denotes the surprisal value derived from a language model parameterized by $\mathbf{\theta}$.
\begin{align}
\Delta P(+\mathbf{Cxn})
&=\iota_{\theta}\!\left(T_{-\mathbf{Plaus.}} \mid S_{+\mathbf{Cxn}}\right) \\ \notag
&\quad - \iota_{\theta}\!\left(T_{+\mathbf{Plaus.}} \mid S_{+\mathbf{Cxn}}\right)
\end{align}


$\Delta P(-\mathbf{Cxn})$ for the control condition with ``or'' is computed similarly. Finally, given a dataset with $k$ example pairs, we then derive an accuracy score by comparing the \pf condition and the control (``or'') condition:
\begin{equation}\label{eq:acc}
   \mathbf{Acc_{PF}} = \frac{1}{k} \sum_{i=1}^{k} \mathds{1}[\Delta P(+\mathbf{Cxn}) >\Delta P(-\mathbf{Cxn})]
\end{equation}
There are other confounding factors (beyond the \pf construction and world knowledge) that could influence LM probability associated with the follow-up sentences. Since the two follow-ups are minimal pairs (with only one word different between them) there is a potential confound of lexical bias if the correct sentence has a more frequent/probable word irrespective of context (e.g.\ since ``easier'' is more frequent than ``harder'' overall). Another potential confound is if there is an ordering effect of the two focused elements, where a follow-up sentence is preferred or dispreferred for presenting focused elements in the same order as the \pf sentence. We control for both of these confounds by balancing both the plausible and implausible follow-ups in our dataset by ordering and by the lexical items that appear in the entailed follow-up. 







\subsection{Syntactic Evaluation}

Beyond evaluating \pf semantics, we also wish to test if models have knowledge of the \textit{forms} of \pf constructions. We create a syntactic evaluation suite which adapts the syntactic tests from \citet{scivetti-etal-2025-unpacking} to our dataset. Specifically, we select three syntactic manipulations from \citet{scivetti-etal-2025-unpacking} which are ungrammatical for \pf constructions but grammatical for simple conjunctions (see \cref{tab:examples_syntax}). Similarly to our \pf semantic tests, we evaluate the $\Delta P$ values between grammatical and ungrammatical sentences compared to simple conjunctions. For more details on the syntactic evaluation suite, as well as evaluations of Global Affinity \citep{rozner-etal-2025-constructions}, see Appendix \ref{sec:appendix_syn1}. 

\begin{table*}[]
\centering\small\smaller\setlength{\tabcolsep}{5pt}
\begin{tabular}{@{}llll@{}}
\toprule
Manipulation        &   Feat. & Manipulated Sentence   & Base  \\
\midrule
Clause Conjunction & \textcolor{purple}{\textbf{+Cxn}}& *I couldn't lift a tiny rock, \textcolor{purple}{\textbf{let alone}} I couldn't lift a huge one. & I couldn't lift a tiny rock, \textcolor{purple}{\textbf{let alone}} a huge one.  \\
Clause Conjunction & \textcolor{yellow}{\textbf{$-$Cxn}}& \hphantom{*}I couldn't lift a tiny rock, \textcolor{yellow}{\textbf{and}} I couldn't lift a huge one. & I couldn't lift a tiny rock, \textcolor{yellow}{\textbf{and}} a huge one.  \\
\midrule
NPI & \textcolor{purple}{\textbf{+Cxn}}& *I could lift a tiny rock, \textcolor{purple}{\textbf{let alone}} a huge one. & I couldn't lift a tiny rock, \textcolor{purple}{\textbf{let alone}} a huge one.  \\
NPI & \textcolor{yellow}{\textbf{$-$Cxn}}& \hphantom{*}I could lift a tiny rock, \textcolor{yellow}{\textbf{and}} a huge one. & I couldn't lift a tiny rock, \textcolor{yellow}{\textbf{and}} a huge one.  \\
\midrule
Pseudocleft & \textcolor{purple}{\textbf{+Cxn}}& *A tiny rock, \textcolor{purple}{\textbf{let alone}} a huge one, I couldn't lift. & I couldn't lift a tiny rock, \textcolor{purple}{\textbf{let alone}} a huge one.  \\
Pseudocleft & \textcolor{yellow}{\textbf{$-$Cxn}}& \hphantom{*}A tiny rock, \textcolor{yellow}{\textbf{and}} a huge one, I couldn't lift. & I couldn't lift a tiny rock, \textcolor{yellow}{\textbf{and}} a huge one.  \\
\bottomrule
\end{tabular}

\caption{\pf syntactic tests. Tests adapted from evaluation paradigm of \citet{scivetti-etal-2025-unpacking}. Accuracy is measured by comparing $\Delta P$ between \textcolor{purple}{\textbf{+Cxn}} and \textcolor{yellow}{\textbf{$-$Cxn}} manipulated sentences versus Base sentences.}
\label{tab:examples_syntax}
\end{table*}

\section{Experiment 1: Impact of Model Size and Training Data}
\label{sec:exp1_main}

\begin{figure}
    \centering
    \includegraphics[width=\columnwidth]{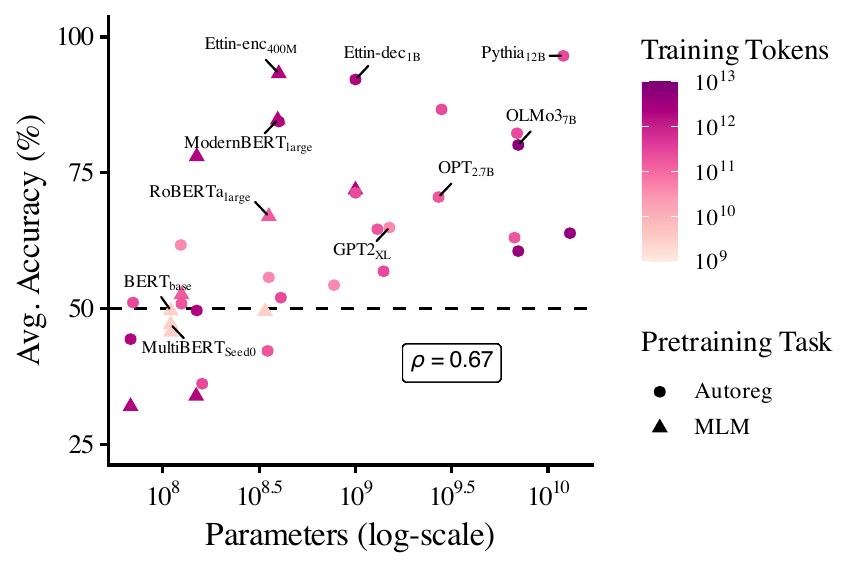}
    \caption{{\pf Semantic Results by Model}. Accuracy is averaged across our four \pf constructions: \la, \ml, \ntm, and \nvm. We observe a high rank-order correlation between model parameters and average accuracy (Spearman's $\rho =0.67$).}
    \label{fig:scaling}
\end{figure}

\subsection{Models and Evaluation}

We test a total of 36 models, varying in parameter count, pretraining data size, pretraining objective (masked vs.~autoregressive)\footnote{For MLM models, we evaluate the probability at the target region (e.g.\ ``easier'' vs.~``harder''), whereas for the decoder models, we evaluate the probability of the entire follow-up sentence conditioned on the first sentence. See Appendix \ref{sec:appendix_sem_mlm} for a replication with MLM scoring using Psuedo-log-likelihood.}, and model family. See Table \ref{tab:models} in Appendix \ref{sec:appendix_models} for model details. In addition to computing accuracy, we run a linear mixed effects model analysis to test the impact of model size, pretraining data, and architecture across all models. The dependent variable is \pf accuracy (see Equation \ref{eq:acc}). We include pretraining data (log \# of tokens), parameters (log \# of parameters), and architecture (causal vs.~MLM) as fixed effects, and a random intercept for model name.

\subsection{Results}

Figure \ref{fig:scaling} visualizes the \pf accuracy\footnote{We average accuracy scores across each of our four \pf constructions.} as a function of model size and training data amount. Below a parameter threshold of roughly 400M, model performance is generally at or below chance regardless of the amount of training data (with the exception of Ettin-encoder-150M). Beyond this parameter threshold, there is a generally positive relationship between training data and accuracy as well as parameter count and accuracy. However, there is substantial variation between individual models, regardless of parameter count, training data, and training objective. 

For our linear modeling analysis, we find that only parameter count has a significant main effect ($\beta= 6.055$, $p=.011$). As a follow-up analysis, we additionally fit a series of linear mixed effects models assessing each predictor independently. For single predictor models (again with a random intercept for model name), we find that both parameter count ($\beta= 6.607$, $p=003$) and pretraining data ($\beta= 2.651$, $p=.034$) are significant, though the effect of parameter count remains larger. While we note that MLM and autoregressive scoring is not directly comparable, we find no significant effect of pretraining objective.  

Regarding our \pf form evaluations, we find that even very small models have strong performance on our syntactic evaluation suite, and assign high Global Affinity scores to fixed slots in the construction (see Tables \ref{tab:syntactic_full} and \ref{tab:aff_results} in Appendix~\ref{sec:appendix_syn1}), echoing past results \citep{rozner-etal-2025-constructions,rozner-etal-2025-babylms,scivetti-etal-2025-unpacking}. The dissociation between tasks for small models underscores the need to evaluate both form and semantics when assessing LMs' overall knowledge of rare constructions.

In summary, these results indicate that models far smaller than frontier LLMs can grasp \pf constructions, as we find that a few models as small as 400M parameters succeed at learning (with 90\%+ accuracy) both the form and semantics of the constructions, and most of the larger models are above chance. There seems to be a broadly positive relationship between parameter count and semantic accuracy, while the effect of training data independent of parameter count is not well supported by our results. Knowledge of the form of \pf constructions is present even in the smallest models we test.


\section{Experiment 2: Learning Trajectory Experiments}

In this experiment, we investigate when \pf constructions are learned in training in relation to other linguistic knowledge. We hypothesize that prior to model knowledge of \pf semantics, we will observe knowledge of constructional \pf forms. We further hypothesize that \pf examples which align with appropriate scalar semantics (given general world knowledge) will be acquired first, especially for models that are less proficient at understanding the constructions.

To operationalize semantic and formal knowledge of \pf constructions, we use \pf accuracy for semantics (Equation~\ref{eq:acc}) and our syntactic test suite identical to Experiment 1. In order to test how alignment of scalar semantics with world knowledge interacts with learning of \pf constructions, we further test models on sentences where constructional examples entail implausible follow-up sentences and contradict plausible sentences (see 10 and 11 in Table \ref{tab:examples}). Like in Experiment 1, we consider an example correct if the presence of the construction shifts increases the probability of the follow-up entailed by the construction, relative to a baseline conjunction ``or''. Intuitively, if a model has a fully abstract understanding of \pf semantics, we expect the presence of a \pf construction to increase the probability of an otherwise implausible statement that is consistent with the scalar relationship implied by the construction.

\begin{figure}
    \centering
    \includegraphics[width=\columnwidth]{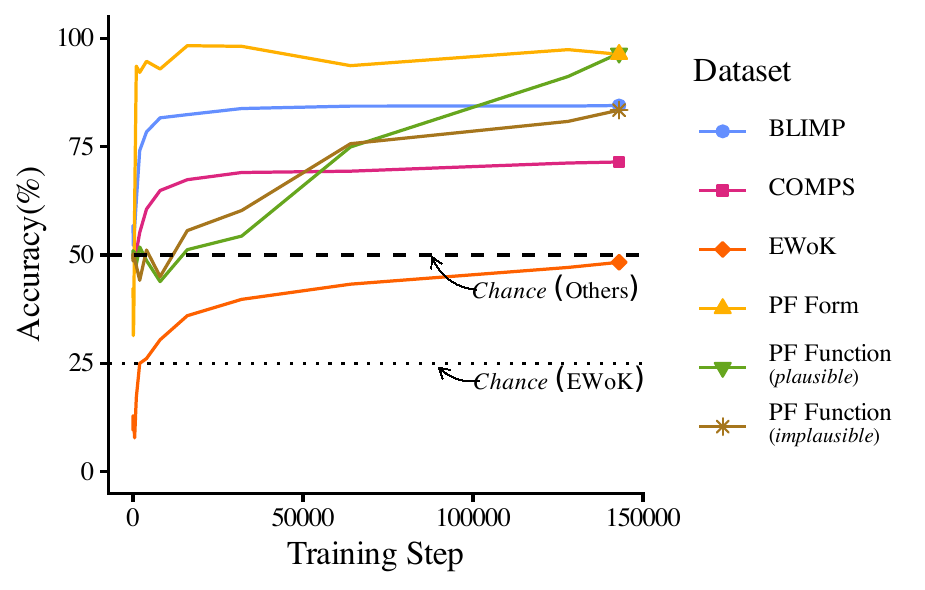}
    \caption{Training dynamics of Pythia-12b on \pf evaluations as well as other linguistic benchmarks. Chance performance on EWoK is 25\%, while chance performance on all other evaluations is 50\%.}
    \label{fig:dynamics1}
\end{figure}

\subsection{Comparison with Linguistic Benchmarks}

We further evaluate the learning dynamics for three other linguistic benchmarks as comparison points throughout training. We evaluate BLiMP \citep{warstadtBLiMPBenchmarkLinguistic2020b}, COMPS \citep{Misra_Rayz_Ettinger_2023}, and EWoK \citep{ivanova-etal-2025-elements}. BLiMP tests model knowledge of a range of syntactic constructions. We expect this dataset to be learned relatively early. COMPS tests knowledge of conceptual properties of noun classes. Knowledge of some properties tested in COMPS may be relevant to linking the noun phrases in our dataset with appropriate scalar semantics, and thus we expect high accuracy on COMPS to precede \pf semantic accuracy. EWoK tests a range of domains of world knowledge, including physical and material properties. Understanding of these properties is crucial to interpreting the scales in our dataset, and thus we expect that high performance on EWoK will correlate with \pf semantic accuracy.

\subsection{Models}

We focus on learning dynamics of three models in Experiment 1: Pythia-12b \citep{Biderman_Schoelkopf_Anthony_Bradley_OBrien_Hallahan_Khan_Purohit_Prashanth_Raff_etal._2023}, Ettin-encoder-400m, and Ettin-decoder-1b \citep{Weller_Ricci_Marone_Chaffin_Lawrie_Durme_2025}. We select these models because they are the top three performing models at their final checkpoints in Experiment 1. For Pythia-12b, we sample 19 logarithmically spaced checkpoints from start to finish in training. For Ettin models, where logarithmically spaced early checkpoints are not available, we evaluate the first 50 checkpoints\footnote{Each checkpoint is equivalent to roughly 8--9 billion tokens of pretraining data, and 50~checkpoints is thus equivalent to approx.\ 400~billion pretraining tokens.} for each model. For each of these model checkpoints, we evaluate \pf semantic accuracy (for both plausible and world-knowledge implausible examples), syntactic accuracy, and accuracy on each of the comparisons. 

\begin{figure}
    \centering
    \includegraphics[width=\columnwidth]{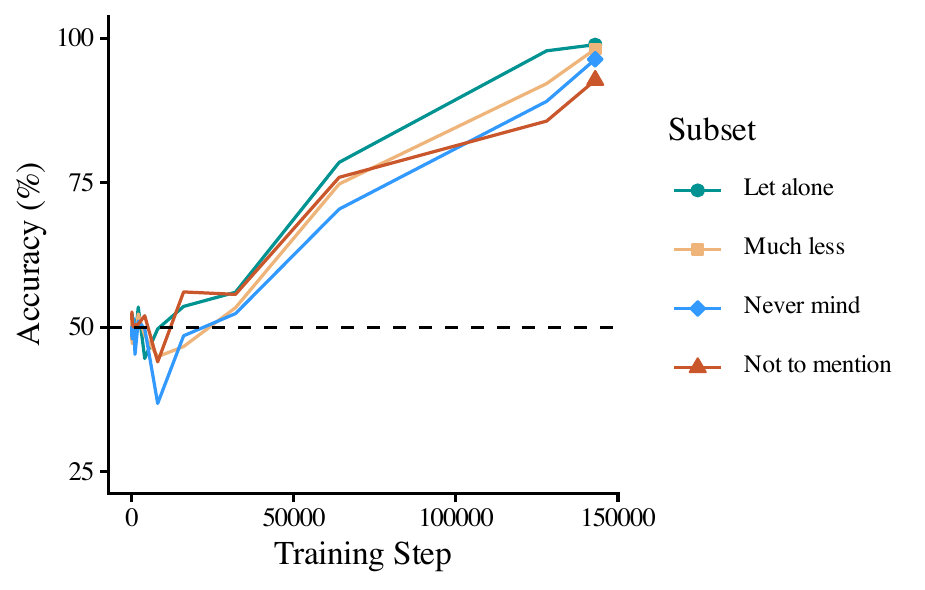}
    \caption{Training dynamics of Pythia-12b for each of the four individual constructions.}
    \label{fig:dynamics1_perconstruction}
\end{figure}

\subsection{Results}


\begin{figure*}[t]
    \centering
    
    \begin{subfigure}[t]{0.32\textwidth}
        \centering
        \includegraphics[width=\linewidth]{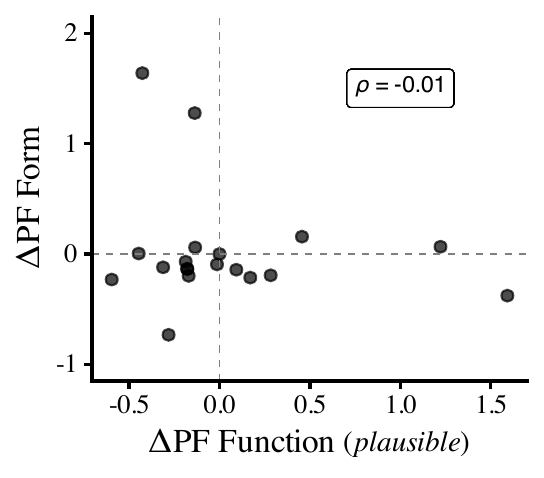}
        \caption{Form vs.~plausible accuracy differences.}
        \label{fig:diffs_scatter2}
    \end{subfigure}
    \hfill
    \begin{subfigure}[t]{0.32\textwidth}
        \centering
        \includegraphics[width=\linewidth]{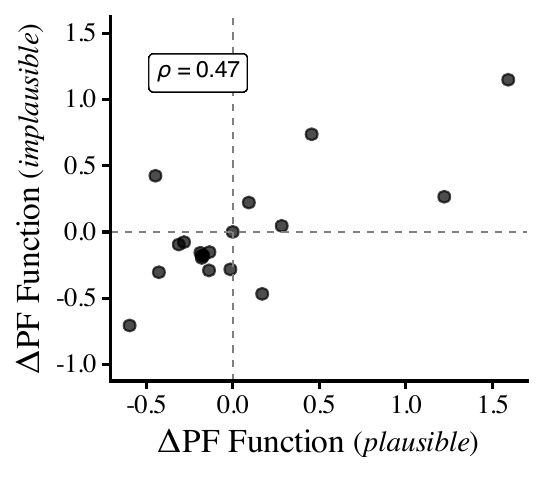}
        \caption{Implausible vs.~plausible accuracy differences.}
        \label{fig:diffs_scatter3}
    \end{subfigure}
    \hfill
    \begin{subfigure}[t]{0.32\textwidth}
        \centering
        \includegraphics[width=\linewidth]{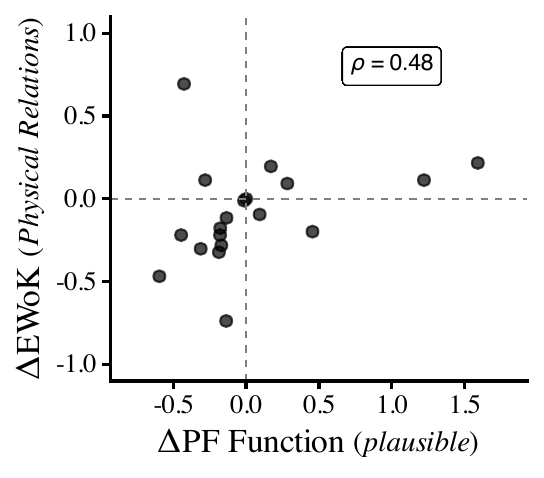}
        \caption{EWoK physical relations vs.~plausible semantic accuracy.}
        \label{fig:diffs_scatter1}
    \end{subfigure}
    
    \caption{
    \textbf{Learning trajectory correlation scatterplots for Pythia-12b.}  Each point represents, for a given checkpoint, how much the model improved over the previous checkpoint with respect to a pair of criteria.
    EWoK physical relations and \pf semantic accuracy show moderate correlation. 
    }
    
    \label{fig:diffs_scatter_combined}
\end{figure*}

For the purpose of clarity, we primarily discuss and visualize results for Pythia-12b in the main body of the paper. Results on Ettin models (for details see Appendix \ref{sec:appendix_exp2}) are broadly qualitatively similar, but display noisier and more inconsistent learning trajectories. We also find that the Ettin models display a more substantial difference in constructional effect on plausible compared to implausible follow-ups, especially early in training. In contrast to the Ettin models, which seem more reliant on world knowledge cues for their interpretations of \pf constructions, Pythia-12b displays a sensitivity to the construction which is robust to implausible contexts. 

Results for Pythia-12b are visualized in Figure~\ref{fig:dynamics1}. We find strong evidence that \pf form is learned prior to \pf semantics. Syntactic accuracy reaches a peak much earlier in training than \pf accuracy. We find performance on \pf semantic evaluations follows similar trajectories regardless of the plausibility of the follow-up sentence, though plausibility does lead to higher absolute performance at the end of training. 

For both BLiMP and COMPS, accuracy plateaus relatively early in training, substantially before the \pf accuracy rises above chance. For EWoK, performance gains are more gradual, similar to gradual gains on \pf semantics. \footnote{Chance accuracy for EWoK is 25\%.}

We report results for each \pf construction individually and graph results for Pythia-12b in Figure \ref{fig:dynamics1_perconstruction}. We observe that performance across the constructions is generally strongly correlated, with all constructions following broadly similar trajectories. Of the four, \la is learned more linearly across training, while performance curves for the other constructions are more logarithmic and peak later in training. While we cannot establish a causal relationship, we note that \la is the most frequent and least ambiguous of the four (see Table \ref{tab:cxn_freqs}) and thus it is perhaps unsurprising that it is learned earlier in training. 




\subsection{Correlation Analysis of Learning Trajectories}


We run a first-difference correlation between \pf semantic performance and performance on other benchmarks. We find negligible correlation between \pf form and semantics, and between \pf semantics and BLiMP. We find a moderate correlation between \pf semantic performance on plausible and implausible follow-ups, further showing Pythia-12b has substantial knowledge of scales implied by \pf constructions beyond the scalar relationships evident through world-knowledge alone. When subdividing EWoK into its different world knowledge domains, we find a moderate correlation between \pf semantics and the physical relations domain ($\rho=.48$). The first-order correlations between \pf formal accuracy, semantic accuracy, and EWoK physical relations accuracy are shown in Figure \ref{fig:diffs_scatter_combined}.  






\subsection{Discussion}

Overall, these results provide strong evidence that \pf form and meaning are learned with vastly different amounts of training input. We find that performance across individual \pf constructions is strongly correlated, and that overall \pf performance is moderately correlated with performance on relevant world knowledge domains in EWoK. We find no evidence that \pf form and meaning acquisition are correlated in any way, nor is \pf semantics significantly correlated with the more syntactic BLiMP benchmark. Regarding model comparison, we find that the learning trajectories of Pythia-12b are much more stable than the trajectories of the smaller Ettin models, which are more susceptible to spikes and valleys in performance across training, and are less proficient at \pf semantics when follow-up sentences are implausible.

\section{Discussion}

In this work, we have shown that medium-sized models ($\approx$400M parameters) can acquire knowledge of \pf semantics. However, small models, and models trained on human-scale data, show a clear gap in performance on \pf form and meaning at their final checkpoints. Furthermore, we have shown that even models which eventually learn \pf meaning only do so much later in training relative to \pf form. The performance gap we observe for \pf constructions echoes past results on \textsc{Comparative-Correlative} \citep{Weissweiler_Hofmann_Köksal_Schütze_2022} and \textsc{Causal-Excess} \citep{zhou-etal-2024-constructions} constructions, which similarly found that LMs fail at semantic evaluations which target those constructions. However, we note that our syntactic tests are inherently different from our semantic tests, and interact with different parts of grammar (e.g., other constructions like conjunction and pseudoclefting for syntactic tests vs.~scalar semantics for semantic tests). While our results do not provide definitive evidence that form and meaning are necessarily learned separately by LMs, they underscore that modeling joint learning of constructional form and meaning (as would be ideally possible for a ``model system'' of constructional acquisition) has not been clearly demonstrated by LMs thus far.

While we do not claim that are results prove that LMs \textit{could never} jointly acquire form and meaning via text-only language modeling, there are obvious reasons why LMs would be unlikely to model a constructionist account of language development. Humans learn language in social settings in order to perform communicative goals. We are also exposed to rich non-linguistic input in the form of embodied experience. Pure text language models do not have access to these rich sources of input, which are likely particularly relevant for extracting semantic and pragmatic features of language. We look towards future work on multimodal human-scale models \citep{hu-etal-2024-findings,Wang_Chandra_Liu_Saligrama_Gong_2025} and models that incorporate human feedback \citep{Ziegler_Stiennon_Wu_Brown_Radford_Amodei_Christiano_Irving_2020} or learning in situated contexts \citep{beuls-van-eecke-2024-humans,botoko-ekila-etal-2025-constructions} as potential avenues for future work.

Finally, we highlight the importance of evaluation design. By improving on past evaluations, we find evidence of \pf semantic learning in models that are much smaller than previously reported. Further innovation in evaluation methods for constructional meaning may yet reveal that functional abilities emerge earlier in training, and at smaller parameter scales, than found here.

\section{Conclusion}

In this work, we have investigated if LMs can learn nuanced semantic interpretations for a rare family of \pf constructions. We find that models larger than $\approx$400M parameters---both encoders and decoders---are broadly successful at the task, with larger models generally performing better. We find that smaller models completely fail at our novel semantic benchmark despite robust knowledge of the constructions' forms. This seeming divergence underscores a broader lack of evidence that text-only LMs are jointly modeling form and meaning of constructions. Turning to an analysis of learning dynamics, we show that learning of \pf semantics occurs after learning formal knowledge of the construction, and is correlated with learning of relevant world knowledge. Our largest high performing model is robust to changes in plausibility when interpreting \pf examples, while smaller models are more sensitive to the plausiblity of the construction. Overall, this work shows that relatively modest-sized models can acquire nontrivial semantic knowledge of rare constructions, and finds correlational evidence relating learning of \pf semantics to learning trajectories of other realms of linguistic knowledge. 

\section*{Limitations}

This paper is limited in terms of its coverage of constructions. While we evaluate a range of \pf constructions in English, it is not clear if our concrete findings would generalize to other constructions which may be less closely related than the set we test here. 
In addition, while we hope that our work serves as a test case of linking the semantics of a set of rare constructions to other semantic properties of language, our evidence in this paper is not causal: while performance on \pf semantics is correlated with certain realms of world knowledge, it is not clear how learning would be impacted if such knowledge were perturbed or removed altogether from training. Furthermore, while models displayed some robustness to implausible follow-up sentences, it is not clear how plausibility is causally linked to learning of \pf constructions. 
The number of scales that we used to create our dataset was somewhat small; increasing the number of scales used (and using less common scales) could impact the results here, especially for smaller models, which display some borderline knowledge of \pf semantics. Finally, \pf constructions are not unique to English, but our evaluations were limited to English constructions and primarily monolingual models. 

        

\section*{Acknowledgments}
Leonie Weissweiler was supported by a postdoctoral fellowship from the German Research Foundation (DFG, \texttt{WE 7627/1-1}). This research was supported in part by NSF award IIS-2144881. We thank CoNLL reviewers and members of the NERT and PiCOL labs for their insightful comments which improved the final version of this paper. 

\bibliography{custom}

\appendix

\section{Dataset Examples}
\label{sec:appendix_data}

This Appendix contains examples for each frame and construction. See Table \ref{tab:frames}. All examples in our evaluations are presented with all four \pf constructions: \la, \ml, \ntm, and \nvm. 

\begin{table*}[]
    \centering\smaller
    \def\arraystretch{1.25}
    \setlength{\tabcolsep}{4pt}
    \begin{tabular}{lrrl}
        \toprule
        \textbf{Scale} & \textbf{\# Num Adjectives} &\textbf{\# Templates} & \textbf{Example} \\
        beautiful --- ugly & 6 & 24 & You couldn't paint an ugly picture, \textcolor{purple}{\textbf{let alone}} a gorgeous one.  \\
        bright --- dim & 2 & 2 & They couldn't see a bright light, \textcolor{purple}{\textbf{let alone}} a dim one. \\
        good --- bad & 8 & 112 & We couldn't cook a bad meal, \textcolor{purple}{\textbf{let alone}} a great one.  \\
        small --- large & 10 & 60 & You couldn't pick up a tiny rock, \textcolor{purple}{\textbf{let alone}} a huge one.  \\
        
    \bottomrule
    \end{tabular}
    \caption{Example \pf sentences for each scale. The number of templates refers to the number of unique combinations of verbs, adjectives, and nouns that were appropriate for that scale. The number of adjectives on the scale is based on \citet{wilkinson-tim-2016-gold}.}
    \label{tab:frames}
\end{table*}

\section{Experiment 1 MLM Replication with Pseudo-Log-Likelihood}
\label{sec:appendix_sem_mlm}

Here, we replicate the semantic tests from Experiment 1 for MLMs using Pseudo-log-likelihood (PLL, \citealp{salazar-etal-2020-masked}), using the formulation from \citet{kauf2023better}. We use the minicons library for computing all PLL scores. Table \ref{tab:pll_results} reports the accuracy on the \pf semantic benchmark as measured by PLL (compare with Table \ref{tab:sem_full}, which contains the full results for the semantic evaluations from Experiment 1). We see similar trends in accuracy as we do in our main evaluation for Experiment 1, where we compare probabilities of masked tokens directly. All MLMs with less than 150 million parameters achieve chance or near chance performance in both settings. We find that ettin-400m, ettin-1b, and ModernBERT-large are the top performing models in both settings.

\begin{table*}[ht]
\centering
\small\smaller
\begin{tabular}{llccccc}
\toprule
\textbf{Architecture} & \textbf{Model} & \la & \ml & \ntm & \nvm & Avg\ \\
\midrule
\multirow{2}{*}{BERT}&base-uncased & 44.4 &50.1 &49.7 &50.4 &48.7 \\
&large-uncased & 32.8 &45.6 &49.2 &43.7 &42.8 \\
\midrule
\multirow{4}{*}{Ettin Encoder }&150m & 76.9 & 83.2 & 68.9 & 75.5 & 76.1\\
&1b & 77.0 & 94.4 & 55.7 & 79.3 & 76.6 \\
&400m & 91.4 & 91.8 & 94.5 & 92.2 & 92.5 \\
&68m & 31.4 & 45.5 &27.6& 27.0& 32.9\\
\midrule
\multirow{2}{*}{ModernBERT}&base & 36.6 & 34.3 &34.0& 41.3& 36.5\\
&large & 87.4 & 93.6& 80.9& 83.6& 86.4 \\
\midrule
\multirow{2}{*}{multiBERTs}&seed\_0 & 47.5 &45.9 &42.7& 48.4& 46.1 \\
&seed\_1 & 44.3 &48.6 &43.1 &42.5 &44.6 \\
\midrule
\multirow{2}{*}{RoBERTa}&base & 50.2 &46.3 &47.2& 51.4 &48.8 \\
&large & 70.6 &65.6 &58.4 &68.6 &65.8 \\
\bottomrule
\end{tabular}

\caption{Semantic Evaluation Scores for MLMs, as scored using Psuedo-log-likelihood \citep{kauf2023better}.}
\label{tab:pll_results}
\end{table*}

\section{Experiment 1: Syntactic Evaluations}
\label{sec:appendix_syn1}

We develop a syntactic evaluation suite which is based upon several tests from \citet{scivetti-etal-2025-unpacking}. The tests focus on three grammatical properties, specifically targeting alternations which are grammatical with simple conjunctions but generally ungrammatical for \pf constructions (see Table \ref{tab:examples_syntax}). Similar to our evaluations on \pf semantics, we measure $\Delta P$ between two conditions, relative to the difference in those conditions when a simple conjunction is present. Unlike the semantic tests, there are no follow-up sentences; rather, we evaluate the differences in probabilities for sentences with \pf constructions directly. More specifically, given a base sentence with a \pf construction $S_{-\mathbf{Manip.,+\mathbf{Cxn}}}$, we expect that an ungrammatically manipulated sentence $S_{+\mathbf{Manip.,\pm\mathbf{Cxn}}}$ will have a relatively high surprisal value. Thus, we compute for a \pf example:

\begin{equation}
\begin{aligned}
\Delta P(+\mathbf{Cxn})
&=\iota_{\theta}\!\left(S_{+\mathbf{Manip.,+\mathbf{Cxn}}} \right) \\
&\quad - \iota_{\theta}\!\left(S_{-\mathbf{Manip.,+\mathbf{Cxn}}}\right)
\end{aligned}
\end{equation}

We then compare to $\Delta P(-\mathbf{Cxn})$, which is computed similarly using ``or'' sentences with manipulations. We calculate accuracy based on $\Delta P$ values identically to the \pf semantics dataset:
\begin{equation}\label{eq:acc2}
   \mathbf{Acc_{PF}} = \frac{1}{k} \sum_{i=1}^{k} \mathds{1}[\Delta P(+\mathbf{Cxn}) >\Delta P(-\mathbf{Cxn})]
\end{equation}

Results on our syntactic evaluation suite are reported in \ref{tab:syntactic_full}.

Additionally, to test general familiarity with the wordforms associated with each construction, we use the \textit{Global Affinity} metric presented by \citet{rozner-etal-2025-constructions}. Global Affinity is defined as the probability assigned to a target word when it is masked in a string. Given an original string $s$ and an index $i$, Global Affinity defines $P_{s \setminus\{i\}}$ as the probability distribution at position $i$. The Global Affinity is then the probability assigned to the correct word $w_i$ when position $i$ is masked:\footnote{Global Affinity is only defined for Masked Language Models \citep{rozner-etal-2025-constructions}. Thus, we only evaluate a subset of our models on this metric.}
\begin{equation}\label{eq:glob_aff}
\mathbf{GlobalAff}_{s,w_i} = P_{s \setminus \{i\}}(w_i)
\end{equation}

In our case, we calculate the Global Affinity for each word in our target \pf constructions (e.g., ``much'' and ``less'' for ``much less'') and average across all words in the construction. We find that regardless of model size, most models have a very high global affinity for all fixed words in \pf constructions (Table \ref{tab:aff_results}). This indicates that models are confident that these \pf constructions are collocations and the fixed words in the constructions are relatively easy to predict for models of all sizes.

\begin{table*}[]
     \resizebox{\textwidth}{!}{
    \centering
\begin{tabular}{lllrrrrr}
    \toprule
    \textbf{Model Type} & \textbf{Family} & \textbf{Model Name} & \la & \ml & \ntm & \nvm & \textbf{Avg.} \\
    \midrule
    \multirow{10}{*}{MLM}
    & \multirow{2}{*}{
  \begin{tabular}[c]{@{}l@{}}
    BERT \\
    \citep{Devlin_Chang_Lee_Toutanova_2019}
  \end{tabular}
}
        & BERT-base-uncased & 48.9	&49.8	&50.1&	49.8&	49.6\\
    & & BERT-large-uncased & 41.0 &59.8	&49.7	&47.6&	49.5\\
    \cmidrule(lr){2-8}
    &\multirow{2}{*}{
  \begin{tabular}[c]{@{}l@{}}
    ModernBERT \\
    \citep{Warner_Chaffin_Clavi_Weller_Hallstrm_Taghadouini_Gallagher_Biswas_Ladhak_Aarsen_et_2024}
  \end{tabular}
}
        & ModernBERT-Base & 34.1 & 32.6 & 31.3 & 37.7 & 33.9\\
    & & ModernBERT-Large & 84.4 & 92.7 & 82.3 & 79.6 & 84.8\\
    \cmidrule(lr){2-8}
    & \multirow{2}{*}{
  \begin{tabular}[c]{@{}l@{}}
    MultiBERT \\
    \citep{Sellam_Yadlowsky_Tenney_Wei_Saphra_DAmour_Linzen_Bastings_Turc_Eisenstein_etal_2021}
  \end{tabular}
}
        & MultiBERT seed 0 & 46.0 & 49.4 & 43.4 & 49.5 & 47.1\\
    & & MultiBERT seed 1 & 48.2 & 48.6 & 44.5 & 41.5 & 45.7\\
    \cmidrule(lr){2-8}
    & \multirow{2}{*}{
  \begin{tabular}[c]{@{}l@{}}
    RoBERTa \\
    \citep{Liu_Ott_Goyal_Du_Joshi_Chen_Levy_Lewis_Zettlemoyer_Stoyanov_2019}
  \end{tabular}
}
        & RoBERTa-base & 51.5 & 49.1 & 56.5 & 53.3 & 52.6\\
    & & RoBERTa-large & 71.7 & 68.8 & 56.8 & 70.7 & 66.9\\
    \cmidrule(lr){2-8}
    & \multirow{4}{*}{
  \begin{tabular}[c]{@{}l@{}}
    Ettin \\
    \citep{Weller_Ricci_Marone_Chaffin_Lawrie_Durme_2025}
  \end{tabular}
}
        & Ettin-Enc-68M & 29.3 & 45.7 & 26.5 & 26.6 & 32.0\\
    & & Ettin-Enc-150M & 81.3 & 84.2 & 69.3 & 77.3 & 78.0\\
    & & Ettin-Enc-400M & 92.0 & 91.8 & 95.9 & 93.4 & 93.3\\
    & & Ettin-Enc-1B & 65.6 & 94.4 & 57.4 & 70.1 & 71.9\\
    \midrule
    \multirow{25}{*}{CausalLM}
    & \multirow{4}{*}{
  \begin{tabular}[c]{@{}l@{}}
    Ettin \\
    \citep{Weller_Ricci_Marone_Chaffin_Lawrie_Durme_2025}
  \end{tabular}
}
        & Ettin-Dec-68M & 44.9 & 41.5 & 44.4 & 46.9 & 44.4\\
    & & Ettin-Dec-150M & 52.3 & 50.9 & 46. & 49.5 & 49.7\\
    & & Ettin-Dec-400M & 88.4 & 83.1 & 84.1 & 82.2 & 84.4\\
    & & Ettin-Dec-1B & 92.3 & 95.4 & 87.9 & 93.1 & 92.2\\
    \cmidrule(lr){2-8}
    & \multirow{4}{*}{
  \begin{tabular}[c]{@{}l@{}}
    GPT2 \\
    \citep{radford2019language}
  \end{tabular}
}
        & GPT2 & 65.4 & 75.3 & 50.1 & 56.0 & 61.7\\
    & & GPT2-medium & 66.3 & 58.2 & 48.6 & 49.9 & 55.7\\
    & & GPT2-large & 74.0 & 54.9 & 37.9 & 50.5 & 54.3\\
    & & GPT2-xl & 74.1 & 65.2 & 57.3 & 63.1& 64.9\\
    \cmidrule(lr){2-8}
    & \multirow{3}{*}{
  \begin{tabular}[c]{@{}l@{}}
    OLMo\\
    \citep{OLMo_Walsh_Soldaini_Groeneveld_Lo_Arora_Bhagia_Gu_Huang_Jordan_etal._2025}\\
    \citep{Olmo_Ettinger_Bertsch_Kuehl_Graham_Heineman_Groeneveld_Brahman_Timbers_Ivison_etal._2026}
  \end{tabular}
}
        & Olmo2-7b & 64.1 & 69.0 & 57.8 & 51.4 & 60.6\\
    & & Olmo2-13b & 73.0 & 70.8 & 60.1 & 51.7 & 63.9\\
    & & Olmo3-7b & 79.9 & 87.4 & 69.2 & 84.0 & 80.1\\ 
    
    \cmidrule(lr){2-8}
    & \multirow{5}{*}{
  \begin{tabular}[c]{@{}l@{}}
    OPT \\
    \citep{zhangOPTOpenPretrained2022}
  \end{tabular}
}
        & OPT-125M & 50.8 & 56.6 & 49.0 & 47.2 & 50.9\\
    & & OPT-350M & 50.6 & 49.7 & 33.2 & 35.5 & 42.2\\
    & & OPT-1.3b & 69.6 & 66.7 & 55.5 & 66.7 & 64.6\\
    & & OPT-2.7b & 72.8 & 69.5 & 70.5 & 69.4 & 70.5\\
    & & OPT-6.7b & 65.9 & 63.9 & 54.7 & 67.9 & 63.1\\
    \cmidrule(lr){2-8}
    & \multirow{8}{*}{
  \begin{tabular}[c]{@{}l@{}}
    Pythia \\
    \citep{Biderman_Schoelkopf_Anthony_Bradley_OBrien_Hallahan_Khan_Purohit_Prashanth_Raff_etal._2023}
  \end{tabular}
}
        & Pythia-70m & 52.7 & 47.9 & 45.5& 58.5& 51.1\\ 
    & & Pythia-160m & 36.2 & 37.2 &  34.9 & 36.3 & 36.2 \\ 
    & & Pythia-410m & 62.9 & 65.0 & 41.4 &38.7 & 52.0\\ 
    & & Pythia-1b & 80.6 & 72.6& 68.3& 63.6&71.3\\ 
    & & Pythia-1.4b & 61.3 & 63.3 & 52.5 & 50.5 & 56.9\\
    & & Pythia-2.8b & 94.4 & 79.1& 83.9& 89.3&86.7\\
    & & Pythia-6.9b & 89.0 & 89.6& 67.4& 83.2&82.3\\
    & & Pythia-12b & 98.9 & 98.1& 92.7& 96.3& 96.5\\
    \bottomrule
\end{tabular}
}
\caption{Full Results on our Semantic Test Suite.}
\label{tab:sem_full}
\end{table*}

\begin{table*}[]
     \resizebox{\textwidth}{!}{
    \centering
\begin{tabular}{lllrrrrr}
    \toprule
    \textbf{Model Type} & \textbf{Family} & \textbf{Model Name} & \la & \ml & \ntm & \nvm & \textbf{Avg.} \\
    \midrule
    \multirow{10}{*}{MLM}
    & \multirow{2}{*}{
  \begin{tabular}[c]{@{}l@{}}
    BERT \\
    \citep{Devlin_Chang_Lee_Toutanova_2019}
  \end{tabular}
}
        & BERT-base-uncased & 89.8 & 96.1 & 93.6 & 60.2 & 84.9\\
    & & BERT-large-uncased & 59.4 & 56.9 & 88.2 & 73.1 & 69.4\\
    \cmidrule(lr){2-8}
    &\multirow{2}{*}{
  \begin{tabular}[c]{@{}l@{}}
    ModernBERT \\
    \citep{Warner_Chaffin_Clavi_Weller_Hallstrm_Taghadouini_Gallagher_Biswas_Ladhak_Aarsen_et_2024}
  \end{tabular}
}
        & ModernBERT-Base & 97.9 & 97.8 & 99.9 & 99.6 & 98.8\\
    & & ModernBERT-Large & 83.0 & 83.3 & 94.0 & 87.1 & 86.9\\
    \cmidrule(lr){2-8}
    & \multirow{2}{*}{
  \begin{tabular}[c]{@{}l@{}}
    MultiBERT \\
    \citep{Sellam_Yadlowsky_Tenney_Wei_Saphra_DAmour_Linzen_Bastings_Turc_Eisenstein_etal_2021}
  \end{tabular}
}
        & MultiBERT seed 0 & 88.0 & 89.5 & 84.8 & 83.9 & 86.6\\
    & & MultiBERT seed 1 & 89.7 & 90.8 & 76.3 & 67.9&81.1\\
    \cmidrule(lr){2-8}
    & \multirow{2}{*}{
  \begin{tabular}[c]{@{}l@{}}
    RoBERTa \\
    \citep{Liu_Ott_Goyal_Du_Joshi_Chen_Levy_Lewis_Zettlemoyer_Stoyanov_2019}
  \end{tabular}
}
        & RoBERTa-base & 93.3 & 93.6 & 99.4 & 93.3 & 94.9\\
    & & RoBERTa-large & 82.8 & 83.8 & 94.6 & 89.6 & 87.7\\
    \cmidrule(lr){2-8}
    & \multirow{4}{*}{
  \begin{tabular}[c]{@{}l@{}}
    Ettin \\
    \citep{Weller_Ricci_Marone_Chaffin_Lawrie_Durme_2025}
  \end{tabular}
}
        & Ettin-Enc-68M & 88.3 & 93.2 & 93.5 & 96.7 & 92.9\\
    & & Ettin-Enc-150M & 93.2 & 92.4 & 98.7 & 94.9 & 94.8\\
    & & Ettin-Enc-400M & 82.5 & 87.0 & 99.3 & 91.4 & 90.0\\
    & & Ettin-Enc-1B & 89.2 & 90.6 & 98.3 & 89.4 & 91.9\\
    \midrule
    \multirow{25}{*}{CausalLM}
    & \multirow{4}{*}{
  \begin{tabular}[c]{@{}l@{}}
    Ettin \\
    \citep{Weller_Ricci_Marone_Chaffin_Lawrie_Durme_2025}
  \end{tabular}
}
        & Ettin-Dec-68M & 100 & 100 & 89.4 & 95.8 & 96.3\\
    & & Ettin-Dec-150M & 99.9 & 99.9 & 90.7 & 97.8 & 97.1\\
    & & Ettin-Dec-400M & 96.5 & 98.9 & 98.4 & 99.5 & 98.4\\
    & & Ettin-Dec-1B & 86.9 & 93.5 & 89.9 & 94.0 & 91.1\\
    \cmidrule(lr){2-8}
    & \multirow{4}{*}{
  \begin{tabular}[c]{@{}l@{}}
    GPT2 \\
    \citep{radford2019language}
  \end{tabular}
}
        & GPT2 & 100 & 100 & 97.2&100 & 99.3\\
    & & GPT2-medium & 98.2 & 98.6 & 94.4& 98.4& 97.4\\
    & & GPT2-large & 99.8 & 99.5 &97.4 & 99.7&99.1\\
    & & GPT2-xl & 99.0 & 97.7 &89.3 & 98.6& 96.2\\
    \cmidrule(lr){2-8}
    & \multirow{3}{*}{
  \begin{tabular}[c]{@{}l@{}}
    OLMo\\
    \citep{OLMo_Walsh_Soldaini_Groeneveld_Lo_Arora_Bhagia_Gu_Huang_Jordan_etal._2025}\\
    \citep{Olmo_Ettinger_Bertsch_Kuehl_Graham_Heineman_Groeneveld_Brahman_Timbers_Ivison_etal._2026}
  \end{tabular}
}
        & Olmo2-7b & 97.0 & 96.5 & 94.9 & 95.4 & 96.0\\
    & & Olmo3-7b & 91.5 & 99.3 & 94.0& 93.7& 94.6\\ \\
    \cmidrule(lr){2-8}
    & \multirow{5}{*}{
  \begin{tabular}[c]{@{}l@{}}
    OPT \\
    \citep{zhangOPTOpenPretrained2022}
  \end{tabular}
}
        & OPT-125M & 100 & 100 & 94.2 & 99.6 & 98.4\\
    & & OPT-350M & 99.9 & 100 & 92.7 & 93.8 & 96.6\\
    & & OPT-1.3b & 97.5 & 98.6 & 94.7 & 96.5 & 96.8\\
    & & OPT-2.7b & 96.0 & 98.2& 96.1 & 94.0 & 96.1\\
    & & OPT-6.7b & 98.3 & 98.9 & 92.8& 97.9& 97.0\\
    \cmidrule(lr){2-8}
    & \multirow{8}{*}{
  \begin{tabular}[c]{@{}l@{}}
    Pythia \\
    \citep{Biderman_Schoelkopf_Anthony_Bradley_OBrien_Hallahan_Khan_Purohit_Prashanth_Raff_etal._2023}
  \end{tabular}
}
        & Pythia-68m & 100 & 100 & 83.3& 99.8& 95.7\\ 
    & & Pythia-160m & 100 & 100 &  94.0 & 98.1 & 98.0\\ 
    & & Pythia-410m & 99.9 & 100 & 97.9 &99.7 & 99.4\\ 
    & & Pythia-1b & 99.8 & 99.7& 91.4& 99.0&97.5\\ 
    & & Pythia-1.4b & 99.9 & 100 &87.3 & 99.8&96.7\\
    & & Pythia-2.8b & 94.8 & 98.7& 93.2& 99.1&96.4\\
    & & Pythia-6.9b & 99.4 & 99.2& 96.2& 99.6&98.6\\
    & & Pythia-12b & 97.3 & 97.6& 92.3& 97.9& 96.3\\
    \bottomrule
\end{tabular}
}
\caption{Full Results on our Syntactic Test Suite. Results for each construction are averaged across the 3 manipulation types (see Table \ref{tab:examples_syntax}). OLMo2-13b was not run for syntactic tests due to compute constraints.}
\label{tab:syntactic_full}
\end{table*}

\begin{figure}
    \centering
    \includegraphics[width=\columnwidth]{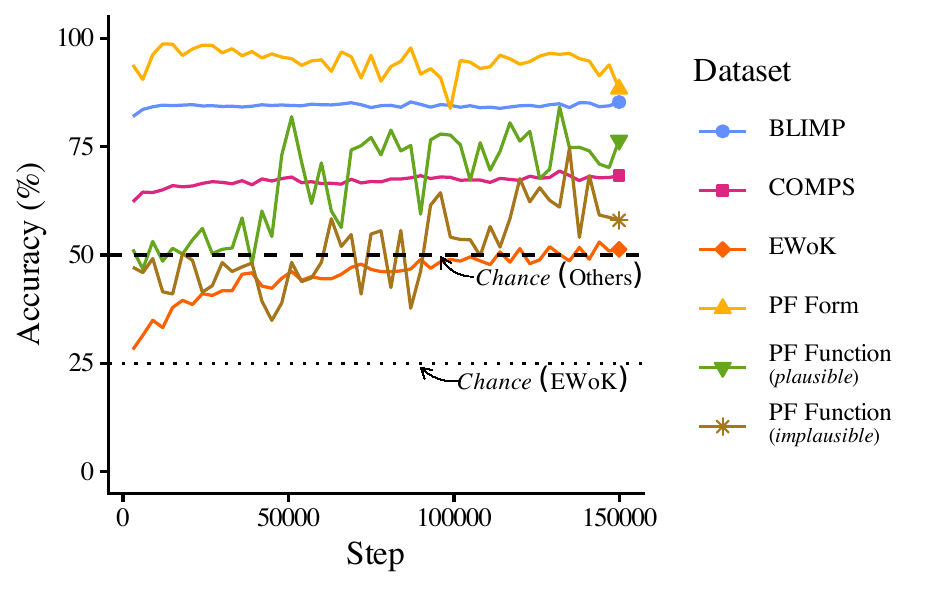}
    \caption{Training dynamics of Ettin-Enc-400M.}
    \label{fig:dynamics2}
\end{figure}

\begin{figure}
    \centering
    \includegraphics[width=\columnwidth]{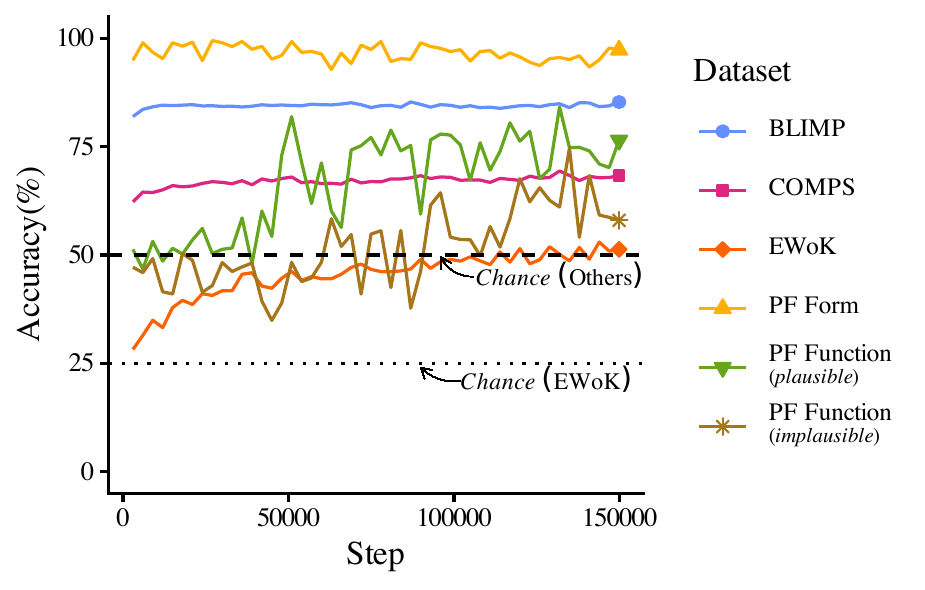}
    \caption{Training dynamics of Ettin-Decoder 1b.}
    \label{fig:dynamics3}
\end{figure}

\begin{table*}[ht]

\centering
\smaller
\begin{tabular}{llcccc}
\toprule
\textbf{Architecture} & \textbf{Model} & \la & \ml & \nvm & \ntm \ \\
\midrule
\multirow{2}{*}{BERT}&base-uncased & $0.999 \pm 0.000$ & $0.987 \pm 0.000$ & $0.502 \pm 0.008$ & $0.933 \pm 0.003$ \\
&large-uncased & $0.999 \pm 0.000$ & $0.999 \pm 0.000$ & $0.835 \pm 0.005$ & $0.999 \pm 0.000$ \\
\midrule
\multirow{4}{*}{Ettin Encoder }&150m & $0.995 \pm 0.000$ & $0.985 \pm 0.000$ & $0.958 \pm 0.001$ & $0.960 \pm 0.001$ \\
&1b & $0.995 \pm 0.000$ & $0.983 \pm 0.000$ & $0.924 \pm 0.001$ & $0.951 \pm 0.001$ \\
&400m & $0.990 \pm 0.000$ & $0.980 \pm 0.000$ & $0.932 \pm 0.001$ & $0.925 \pm 0.001$ \\
&68m & $0.998 \pm 0.000$ & $0.985 \pm 0.000$ & $0.779 \pm 0.003$ & $0.797 \pm 0.005$ \\
\midrule
\multirow{2}{*}{ModernBERT}&base & $0.987 \pm 0.000$ & $0.992 \pm 0.000$ & $0.903 \pm 0.002$ & $0.791 \pm 0.004$ \\
&large & $0.992 \pm 0.000$ & $0.985 \pm 0.000$ & $0.926 \pm 0.002$ & $0.923 \pm 0.002$ \\
\midrule
\multirow{2}{*}{multiBERTs}&seed\_0 & $0.999 \pm 0.000$ & $0.999 \pm 0.000$ & $0.418 \pm 0.007$ & $0.741 \pm 0.006$ \\
&seed\_1 & $0.999 \pm 0.000$ & $0.995 \pm 0.000$ & $0.439 \pm 0.007$ & $0.609 \pm 0.007$ \\
\midrule
\multirow{2}{*}{RoBERTa}&base & $0.998 \pm 0.000$ & $0.998 \pm 0.000$ & $0.944 \pm 0.001$ & $0.998 \pm 0.000$ \\
&large & $0.996 \pm 0.000$ & $0.984 \pm 0.000$ & $0.963 \pm 0.001$ & $0.991 \pm 0.000$ \\
\bottomrule
\end{tabular}

\caption{\textbf{Global Affinity Scores By \pf Construction.} Intervals represent $\pm$ 95\% confidence intervals. Results are only reported for MLMs as \citet{rozner-etal-2025-constructions} do not provide a definition of Global Affinity for CausalLMs.}
\label{tab:aff_results}
\end{table*}

\section{Experiment 2 Full Results}
\label{sec:appendix_exp2}

In this section, we provide extended results on learning dynamics for Ettin-encoder400m and Ettin-decoder1b. Results are shown in Figures \ref{fig:dynamics2} and \ref{fig:dynamics3} respectively. In general, the performance trajectories for \pf semantics are substantially noisier in these models than in Pythia, with large peaks and valleys throughout training. For both Ettin models, we observe early spikes in performance on plausible examples that are not accompanied by spikes on implausible examples. Generally, performance on implausible examples does not rise consistently above chance until much later in training. Taken together, these results seem to indicate that the smaller Ettin models do learn nontrivial \pf semantics, but may be limited to more natural contexts where the scalar relationship entailed by the construction is further supported by world knowledge. Performance on other benchmarks is generally more stable than on \pf semantics, though performance on EWoK is noticeably less stable relative to Pythia.

\section{Model Details}
\label{sec:appendix_models}

Table \ref{tab:models} presents details about the models that we test in Experiment 1. In total, we test 36 models in Experiment~1. In Experiment~2, we test 3~models: Pythia-12b, Ettin-encoder400m, and Ettin-decoder1b, which are selected due to their strong performance in Experiment 1 and the availability of their intermediate checkpoints. 

\begin{table*}[]
    \centering\small\smaller
\begin{tabular}{lllrr}
    \toprule
    \textbf{Model Type} & \textbf{Family} & \textbf{Model Name} & \textbf{Parameter Count} & \textbf{Pretraining Data (\# of Tokens)} \\
    \midrule
    \multirow{10}{*}{MLM}
    & \multirow{2}{*}{
      \begin{tabular}[c]{@{}l@{}}
        BERT \\
        \citep{Devlin_Chang_Lee_Toutanova_2019}
      \end{tabular}
    }
        & BERT-base-uncased & 110M & 3B\\
    & & BERT-large-uncased & 335M & 3B\\
    \cmidrule(lr){2-5}
    & \multirow{2}{*}{
      \begin{tabular}[c]{@{}l@{}}
        ModernBERT \\
        \citep{Warner_Chaffin_Clavi_Weller_Hallstrm_Taghadouini_Gallagher_Biswas_Ladhak_Aarsen_et_2024}
      \end{tabular}
    }
        & ModernBERT-Base & 110M & 2T\\
    & & ModernBERT-Large & 375M & 2T\\
    \cmidrule(lr){2-5}
    & \multirow{2}{*}{
      \begin{tabular}[c]{@{}l@{}}
        MultiBERT \\
        \citep{Sellam_Yadlowsky_Tenney_Wei_Saphra_DAmour_Linzen_Bastings_Turc_Eisenstein_etal_2021}
      \end{tabular}
    }
        & MultiBERT seed 0 & 110M & 3B\\
    & & MultiBERT seed 1 & 110M & 3B\\
    \cmidrule(lr){2-5}
    & \multirow{2}{*}{
      \begin{tabular}[c]{@{}l@{}}
        RoBERTa \\
        \citep{Liu_Ott_Goyal_Du_Joshi_Chen_Levy_Lewis_Zettlemoyer_Stoyanov_2019}
      \end{tabular}
    }
        & RoBERTa-base & 125M & 160B\\
    & & RoBERTa-large & 355M & 160B\\
    \cmidrule(lr){2-5}
    & \multirow{4}{*}{
      \begin{tabular}[c]{@{}l@{}}
        Ettin \\
        \citep{Weller_Ricci_Marone_Chaffin_Lawrie_Durme_2025}
      \end{tabular}
    }
        & Ettin-Enc-68M & 68M & 2T\\
    & & Ettin-Enc-150M & 150M & 2T\\
    & & Ettin-Enc-400M & 400M & 2T\\
    & & Ettin-Enc-1B & 1B & 2T\\
    \midrule
    \multirow{25}{*}{CausalLM}
    & \multirow{4}{*}{
      \begin{tabular}[c]{@{}l@{}}
        Ettin \\
        \citep{Weller_Ricci_Marone_Chaffin_Lawrie_Durme_2025}
      \end{tabular}
    }
        & Ettin-Dec-68M & 68M & 2T\\
    & & Ettin-Dec-150M & 150M & 2T\\
    & & Ettin-Dec-400M & 400M & 2T\\
    & & Ettin-Dec-1B & 1B & 2T\\
    \cmidrule(lr){2-5}
    & \multirow{4}{*}{
      \begin{tabular}[c]{@{}l@{}}
        GPT2 \\
        \citep{radford2019language}
      \end{tabular}
    }
        & GPT2 & 125M & 160B\\
    & & GPT2-medium & 355M & 160B\\
    & & GPT2-large & 775M & 160B\\
    & & GPT2-xl & 1.5B & 160B\\
    \cmidrule(lr){2-5}
    & \multirow{3}{*}{
      \begin{tabular}[c]{@{}l@{}}
        OLMo \\
        \citep{OLMo_Walsh_Soldaini_Groeneveld_Lo_Arora_Bhagia_Gu_Huang_Jordan_etal._2025}\\
        \citep{Olmo_Ettinger_Bertsch_Kuehl_Graham_Heineman_Groeneveld_Brahman_Timbers_Ivison_etal._2026}
      \end{tabular}
    }
        & Olmo2-7b & 7B & 4T\\
    & & Olmo2-13b & 13B & 5T\\
    & & Olmo3-7b & 7B & 6T\\
    \cmidrule(lr){2-5}
    & \multirow{5}{*}{
      \begin{tabular}[c]{@{}l@{}}
        OPT \\
        \citep{zhangOPTOpenPretrained2022}
      \end{tabular}
    }
        & OPT-125M & 125M & 180B\\
    & & OPT-350M & 350M & 180B\\
    & & OPT-1.3b & 1.3B & 180B\\
    & & OPT-2.7b & 2.7B & 180B\\
    & & OPT-6.7b & 6.7B & 180B\\
    \cmidrule(lr){2-5}
    & \multirow{8}{*}{
      \begin{tabular}[c]{@{}l@{}}
        Pythia \\
        \citep{Biderman_Schoelkopf_Anthony_Bradley_OBrien_Hallahan_Khan_Purohit_Prashanth_Raff_etal._2023}
      \end{tabular}
    }
        & Pythia-70m & 70M & 260B\\
    & & Pythia-160m & 160M & 260B\\
    & & Pythia-410m & 410M & 260B\\
    & & Pythia-1b & 1B & 260B\\
    & & Pythia-1.4b & 1.4B & 260B\\
    & & Pythia-2.8b & 2.8B & 260B\\
    & & Pythia-6.9b & 6.9B & 260B\\
    & & Pythia-12b & 12B & 260B\\
    \bottomrule
\end{tabular}
    \caption{Information on all models tested in Experiment 1. Parameter Count and \# of Pretraining Tokens are approximate measures.}
    \label{tab:models}
\end{table*}

\end{document}